\documentclass[letterpaper, 10 pt, conference]{ieeeconf}  

\IEEEoverridecommandlockouts                              

\usepackage{graphics} 
\usepackage{epsfig} 
\usepackage{mathptmx} 
\usepackage{mathtools}
\usepackage{times} 
\usepackage{amsmath} 
\usepackage{amssymb}  
\usepackage[acronym]{glossaries} 
\usepackage{hyperref}
\usepackage{graphicx}
\usepackage{subfigure}
\usepackage[usenames,dvipsnames]{color} 

\newacronym{rl}{RL}{Reinforcement Learning}
\newacronym{drl}{DRL}{Deep Reinforcement Learning}
\newacronym{mdp}{MDP}{Markov Decision Process}
\newacronym{ppo}{PPO}{Proximal Policy Optimization}
\newacronym{sb3}{SB3}{Stable Baselines 3}

\title{\LARGE \bf
Learning Dexterous Object Handover
}

\author{Daniel Frau-Alfaro$^{1}$, Julio Castaño-Amoros$^{1}$, Santiago Puente$^{1}$, Pablo Gil$^{1}$ and Roberto Calandra$^{2}$
\thanks{*This work was partially supported by the Interreg-VI Sudoe and European Regional Development Fund through the \href{https://interreg-sudoe.eu/proyecto-interreg/remain/}{REMAIN Project} under Grant S1/1.1/E0111, by the project ``Genius Robot'' (01IS24083) BMBF, by the DFG as part of EXC 2050/1 – Project ID 390696704 CeTI of Technische Universität Dresden by BMBF, and by the DAAD in project 57616814 (\href{https://secai.org/}{SECAI} \href{https://secai.org/}{School of Embedded and Composite AI}) and by the Spanish Government through the Grant PID2021-122685OB-I00 and by the University of Alicante under Grant UAFPU21-26.}
\thanks{*This work was conducted during Julio's research stay at LASR Lab.}
\thanks{$^{1}$Daniel Frau-Alfaro, $^{1}$Julio Castaño-Amoros, $^{1}$Santiago Puente and $^{1}$Pablo Gil are with the AUROVA Lab, Department of Physics, Systems Engineering, and Signal Theory, University of Alicante, 03690 Alicante, Spain 
        {\tt\small daniel.frau@ua.es}, {\tt\small julio.ca@ua.es}, {\tt\small santiago.puente@ua.es}, {\tt\small pablo.gil@ua.es}}%
\thanks{$^{2}$Roberto Calandra is with LASR Lab, Technische Universitat Dresden, Dresden, Germany
        {\tt\small rcalandra@lasr.org}}%
}

\begin{document}

\maketitle
\thispagestyle{empty}
\pagestyle{empty}

\begin{abstract}

Object handover is an important skill that we use daily when interacting with other humans.
To deploy robots in collaborative setting, like houses, being able to receive and handing over objects safely and efficiently becomes a crucial skill.
In this work, we demonstrate the use of \gls{rl} for dexterous object handover between two multi-finger hands.
Key to this task is the use of a novel reward function based on dual quaternions to minimize the rotation distance, which outperforms other rotation representations such as Euler and rotation matrices.
The robustness of the trained policy is experimentally evaluated by testing w.r.t. objects that are not included in the training distribution, and perturbations during the handover process. 
The results demonstrate that the trained policy successfully perform this task, achieving a total success rate of 94\% in the best-case scenario after 100 experiments, thereby showing the robustness of our policy with novel objects. In addition, the best-case performance of the trained policy decreases by only 13.8\% when the other robot moves during the handover, proving that our policy is also robust to this type of perturbation, which is common in real-world object handovers. Code and videos can be found  \href{https://aurova-projects.github.io/bimanual_handover/}{here}.

\end{abstract}

\section{Introduction}
With the recent focus on humanoid robots, service robots, and human-robot collaboration, several efforts have been made to teach robots how to perform dexterous manipulation tasks, such as collaborative assembly, package manipulation in logistics, and household chores. However, there are still many open questions about how to approach this type of tasks using robots, which require a variety of skills, including a high degree of dexterity, perception, coordination, collaboration, and understanding \cite{survey_object_handover} \cite{DUAN2024100145}. In contrast, humans have an innate talent for performing tasks that require these types of skills. Therefore, it makes sense to bring humans into the loop when tackling these complex problems.

\begin{figure}
    \centering
    \includegraphics[width=1\linewidth]{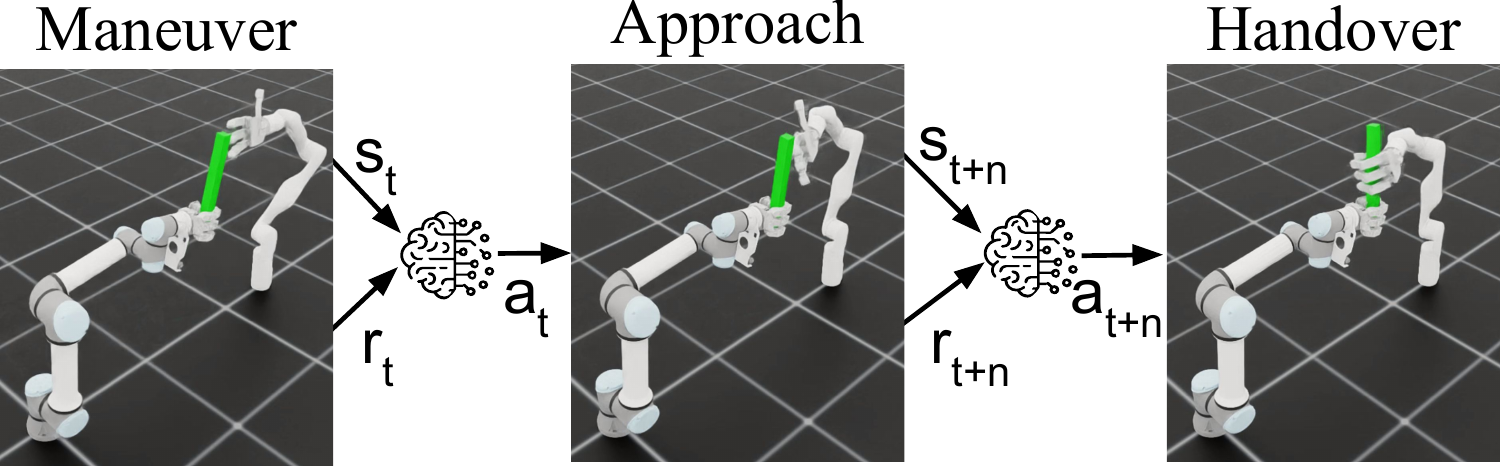}
    \caption{Overview of the different phases of the object handover process. Both robots must collaborate to successfully complete object handover, which is more complicated when employing multi-finger robotic hands. }
    \label{fig:teaser}
\end{figure}

In particular, this work addresses the task of object handover which can be very interesting in the context of human-robot collaboration for both industrial and social applications. According to \cite{intro_human_robot_colab}, the object handover collaboration can be classified into robot-robot collaboration, human-robot collaboration, and robot-human collaboration. The differences between them are whether humans are involved in the task or not, and which participant acts as the giver and which one as the receiver. This type of collaboration typically comprises different subtasks, from the maneuver to set the initial hand pose to the object handover and its subsequent manipulation, as illustrated in Fig.~\ref{fig:teaser}, that must be resolved in order to complete the task.

In this work, we explore the object handover task based on robot-robot collaboration, as an initial step for human-robot and robot-human collaboration. 
We assume that the robotic hand holding the object is already fixed in an arbitrary handover pose. 
Consequently, our objective is to train a single \gls{rl} policy so that a robotic arm and hand can learn how to approach, grasp, and transport the object from another robotic system. The main contributions of this work are twofold. First, in contrast of using 2-finger small grippers as in \cite{rel_work_bim_hand_2}, we use 4-finger articulated hands, which increases the DoF and consequently the difficulty of applying learning-based strategies. In addition, unlike other works such as \cite{bimanual_1} \cite{10657347}, we discard the use of teleoperation systems to teach robots how to perform the task. Second, \cite{trabajo_prev_dani} proposed a dual quaternion-based reward function to avoid rotation constraints in $S0(3)$. The goal of this work is to expand that reward function to the object handover task using 4-finger robotic hands, and to compare it with other representations, such as Euler or matrix rotation, that do require these restrictions.

\section{Related Work}
\label{sec:rel_work}

Traditionally, collaboration tasks such as object handover have been addressed employing control theory algorithms, sensor fusion techniques, probabilistic methods, etc., as in \cite{intro_human_robot_colab}, \cite{intro_human_robot_colab2}, and \cite{intro_human_robot_colab3}. Although the results of such approaches are promising, they are limited for several reasons. For instance, sensor fusion techniques tend to accumulate errors when calibrating different devices. More specifically, the use of control theory algorithms require the design of the policy for each subtask, while our goal is to learn it using \gls{rl}.

In the literature of \gls{rl}-based object handover, this task has been addressed in different ways. 
For instance, \cite{rel_work_bim_hand_1} formulated the object handover task by throwing and catching small objects, while our approach is based on a direct handover using longer objects due to the size of the hands. The authors employed two 6-DoF robotic arms and two Allegro hands, similarly as we do in our work. They proposed a multi-agent approach and a three-phase training, while our policy comprises a single agent and trains on a single-phase process, thereby reducing the number of hyperparameters to tune.
Their trained policies obtained a throw and catch success rate of 95\% and 37\% when testing with 11 known objects and 14 novel objects in simulation, respectively. These results showed a large gap in robustness when evaluating with objects out of the training distribution. Our policy is more robust in this type of case, although we use fewer objects to evaluate in our experiments. 

Although throwing and catching can be considered a type of object handover, it does not involve physical interaction between the robots and the object at the same time, which is what we are interested in exploring in this work. In this context, repositioning objects using two robotic systems requires transferring the object between them, resulting in a direct handover \cite{rel_work_bim_hand_2}. To achieve this, a single-agent policy was trained via actor-critic learning, which is similar to our approach. In this case, the authors employed two 7-DoF robotic arms equipped with two 2-finger grippers to handover objects similar to ours. However, they simplified the task by using grippers instead of multi-finger hands, which require orientation optimization in $SO(3)$, as we propose in our work. They reported an average success rate of 94\% in the simulated environment, which is similar to the results obtained in our experiments, taking into account the complexity of using multi-finger hands with different objects in our approach. 

Humanoid robots equipped with multi-finger hands have also been employed to approach this task using \gls{rl} recently \cite{rel_work_bim_hand_3}. Specifically, two 7-DoF robotic arms and two 5-finger grippers were involved in the robotic setup. Initially, they recorded demonstrations to capture the human pose of the hands at the beginning of the episode, while we let our policy learn how to maneuver to approach the object correctly, which can sometimes produce more optimal results.  As visual input, they performed an ablation study to evaluate the use of the 3D object position together with the depth image. The results of this study reported that their policy was not able to learn with the depth image alone, but with the 3D object position, as we do in our work. Similar to our work, they designed a contact-based reward function. However, they generated several contact markers to guide the policy to grasp the object, while we only define a single grasping frame and let the policy learn how to grasp the object and keep it stable. The main difference of our approach is the use of dual quaternions to learn the orientation of the hands, which is not considered in any of the aforementioned works. 


\begin{figure*}
    \centering
    \includegraphics[width=0.7\linewidth]{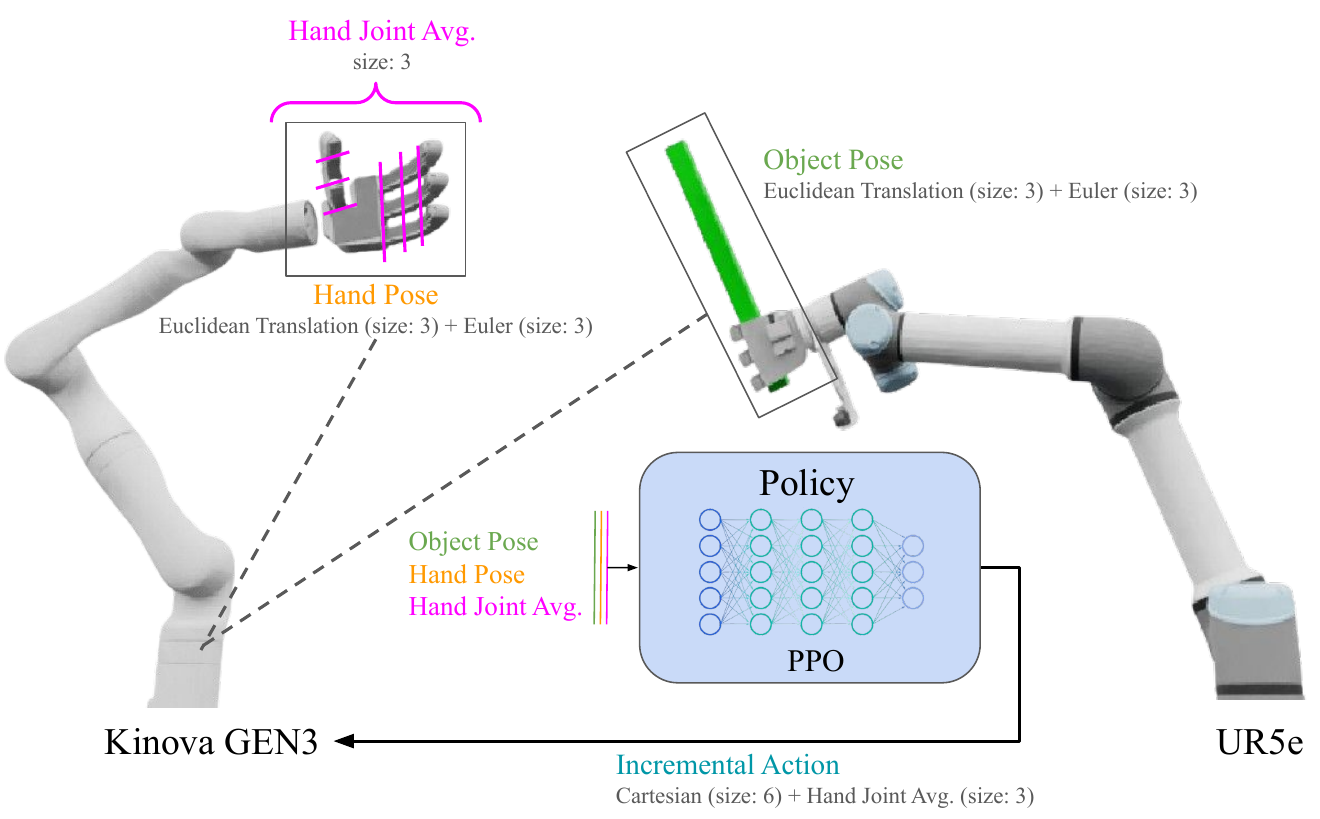}
    \caption{Overview of the \gls{rl} environment used in this work, comprising a single PPO agent. The observations consist of the Euclidean translations and Euler rotations of the GEN3 robot and the object, as well as the average of each joint value for the different fingers, resulting in three global joint values for the whole hand. The policy outputs increments in translation and Euler rotation as actions for the GEN3 end effector, and the three corresponding increment joint values for the hand.}
    \label{fig:env_setup}
\end{figure*}

\section{Mathematical Foundations}
\label{sec:math_foundations}

\gls{rl} and \gls{drl} algorithms allow to obtain policies for a vast range of tasks with high dimensionality of actions and observations, like object handover, which are difficult to deal with classical control techniques.
In this section, a brief introduction to the basics of the \gls{rl} paradigm is provided. Next, we explain the dual quaternion algebra used in this work.

\subsection{RL basics}
The \gls{rl} approach considers every problem as a \gls{mdp} \cite{li2017deep} \cite{wiering2012reinforcement}. Following this modeling, an agent placed in an environment captures information as a state $s_t$. Then, it produces an action $a_t$ based on a certain policy $a_t \sim \pi(\cdot|s_t)$. This produces a change (or step) in the environment, leading to $s_{t+1}$. In addition, the agent receives a reward $r_t$ that indicates how good the action was according to the task it is performing. With this, the final objective of the agent is to generate a policy $\pi$ that maximizes the expected return over time, estimated using \eqref{eq:est_ret}.

\begin{equation} \label{eq:est_ret}
    E[\sum_{t= 0}^T \gamma^t r_t]\,
\end{equation}

where $\gamma \in [0, 1]$ is the discounting factor over time.

\subsection{Dual quaternion algebra}

For the sake of clarity, a brief introduction to dual quaternion algebra is presented  \cite{jia2013dual}  \cite{thomas2014approaching}. 
In the upcoming explanation, we assume some previous knowledge in the field of simple quaternions $\mathbf{q} \in \mathbb{H}$ to represent rotations in $SO(3)$.

Dual quaternions are an extension of the group of dual numbers $\hat q \in \mathcal{H} \; \text{with} \ \hat q = (a + \epsilon b) ,\  (a,b) \in \mathbb{R}$ and with $\epsilon$ being the dual operator that fulfils that $\epsilon^2=0, \; \epsilon \neq 0$. The elements of a dual quaternion are simple quaternions instead of real numbers, therefore, considering $\mathbf{\hat q}$ as a dual quaternion, $\mathbf{\hat q} \in \mathbb{\underline H} \   \text{with } \mathbf{\hat q} = (\mathbf{a} + \epsilon \mathbf{b}) , \; (\mathbf{a}, \mathbf{b}) \in \mathbb{H}$. This formulation allows performing the mathematical operations in Table \ref{tab:dq_operations} given $\mathbf{\hat q_1} = \mathbf{q_{p1}} + \epsilon \mathbf{q_{d1}}$ and $\mathbf{\hat q_2} = \mathbf{q_{p2}} + \epsilon \mathbf{q_{d2}}$ and with $\ll\cdot\gg$ being the dot operation of two vectors.

\begin{table}[t]
\caption{
Basic operations with dual quaternions. We used these operations to calculate the reward function in Section \ref{sec:task_mod}.}\label{tab:dq_operations}
\centering
\renewcommand{\arraystretch}{1.5}
\begin{tabular}{ll}
\hline
\textbf{Operation}                             & \textbf{Formulation} \\ \hline
\textbf{Primary part}       & $\mathcal{P}$($\hat{\mathbf{q}}_1$) $= \mathbf{q_p}_{1}$   \\
\textbf{Dual Part}  & $\mathcal{D}$($\hat{\mathbf{q}}_1$) $= \mathbf{q_d}_{1}$     \\
\textbf{Addition}             & $\hat{\mathbf{q}}_1 + \hat{\mathbf{q}}_2 = \mathbf{q_p}_1 + \mathbf{q_p}_2 + \epsilon(\mathbf{q_d}_1 + \mathbf{q_d}_2)$                                                     \\
\textbf{Multiplication}   & $\hat{\mathbf{q}}_1 \otimes \hat{\mathbf{q}}_2 = \mathbf{q_p}_1 \cdot \mathbf{q_p}_2 + \epsilon(\mathbf{q_p}_1 \cdot \mathbf{q_d}_2 + \mathbf{q_d}_1 \cdot \mathbf{q_p}_2)$ \\
\textbf{Conjugate}        & $\hat{\mathbf{q}}_1^* = \mathbf{q^*_p}_1 + \epsilon \mathbf{q^*_d}_1$ \\
\textbf{Magnitude}           & $||\hat{\mathbf{q}}_1|| = \hat{\mathbf{q}}_1 \otimes \hat{\mathbf{q}}_1^*$    \\
\textbf{Difference}       & $\hat{\mathbf{q}}_{\text{diff}} = \hat{\mathbf{q}}_1^{*} \otimes \hat{\mathbf{q}}_2$   \\ 
\textbf{Identity Element}  & $\hat{\mathbf{I}} = \mathbf{I} + \epsilon 0, \;\; \mathbf{I} = (1 + 0i+0j+0k) \; \in \mathbb{H}$ \\ \hline
\end{tabular}

\end{table}

Dual quaternions are often used to represent poses in $SE(3)$ space, combining translations and rotations in a compact and short formulation. Hence, given a rotation quaternion $\mathbf{q_r}$ and a translation $\mathbf{q_t}$ expressed as a pure quaternion, that is, a quaternion with null real part, the dual quaternion expressing that transformation is written following \eqref{eq:dq_from_tr}.

\begin{equation} \label{eq:dq_from_tr}
    \hat{\mathbf{q}} = \mathbf{q_r} + \epsilon \left(\frac{1}{2} \mathbf{q_r} \cdot \mathbf{q_t}\right)\,,
\end{equation}

where the resulting dual quaternion meets the unitary condition as $||\hat{\mathbf{q}}|| = 1$. The primary part of a dual quaternion represents the rotation of a pose, while the dual part contains information about the translation along that orientation.

There are other formulations so as to represent poses in 3D space. Most of them utilize the Euclidean translation to encode translations and vary the way in which we can express orientations. An example of this issue are homogeneous transformation matrices $SO(3) \rtimes \mathbb{R}^3$. This approach employs 16 values to encode a pose in space, which is less computationally efficient to operate with, compared to the dual quaternion representation.
In addition, we can compute the difference or distance between poses separately in rotation and translation, which requires normalization to combine them. Other representations use translation vectors along with Euler angles to represent positions and rotations $\mathbb{R}^3 \rtimes \mathbb{R}^3$, respectively. However, this alternative also suffers from the separation of distances and other problems, e.g. gimbal lock or singularities when interpolating.

Summarizing, the dual quaternion allows for a unified representation of poses; both translations and rotations are expressed under the same formulation, and they do not need further processing or constraints to compute distances.

\section{Handover System and Reward Function Design}
\label{sec:methodology}

We now introduce the task modeling and the design of our reward function. We define the states and actions in the \gls{rl} framework, as well as the division of the handover task into movement primitives that enable the formulation of a reward function by parts.

\subsection{Object Handover Setup}
In terms of modeling of the environment, we use an UR5e (6 DoF) and GEN3 (7 DoF), both equipped with Allegro Hands and touch sensors to perform the manipulation.
At the start of each episode, we initialize the UR5e to an arbitrary pose chosen randomly. The UR5e robot holds the object without moving during the whole episode as shown in Fig.~\ref{fig:env_setup}. This way, it mimics a human giver which should not be trained. Then, the GEN3 has to learn to reach the object and grasp it from the same starting pose in each episode. The policy takes as observations the Cartesian poses of the GEN3 and the object along with its hand joint values. In terms of actions, we add increments to the current pose of the GEN3 end effector and to the joint values of the hand.

\subsection{Task Modeling} \label{sec:task_mod}
As for the task modeling and the reward function, it is worth mentioning that we divide the object handover task into different movement primitives or phases, as shown in Fig.~\ref{fig:teaser}. In each stage, we change a simple initial reward function defined in \eqref{eq:simple_rew} using different modifiers. These provide the reward function 
\begin{equation} \label{eq:simple_rew}
    r_{\text{BASE}_t} = \eta_t \, e^{-d_t},
\end{equation}
for each phase of the task, where $r_{\text{BASE}_t}$ is a basic reward for approaching the object, $d_t$ is the distance between the pose of the hand and the object at step $t$ and $\eta_t$ is the weight of the reward at step $t$ which will be defined in \eqref{eq:mod_scale}.

\subsubsection{Maneuver phase}
\begin{figure}
    \centering
    \includegraphics[width=0.47\linewidth]{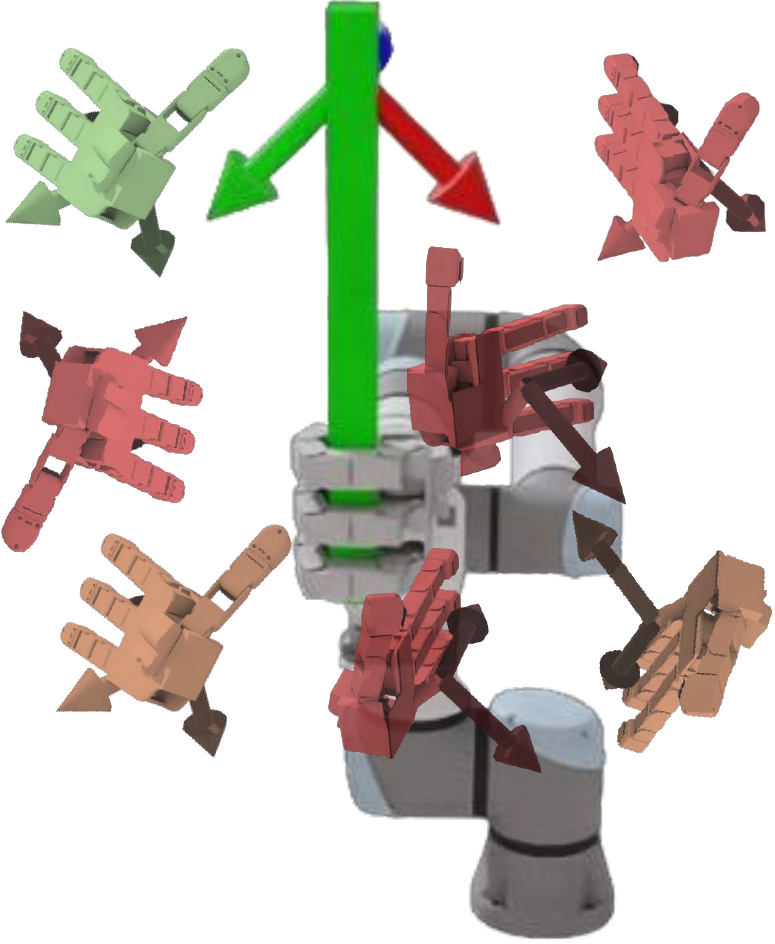}
    \caption{Example of different poses of the GEN3 hand evaluated with respect to the relative pose of the object. The different colors of the GEN3 hand indicate how adequate is the approach to grasp the object correctly. \textbf{Red}: oriented with the back of the hand toward the object, \textbf{Orange}: closer to the UR5e hand than to the object frame, \textbf{Green}: palm facing the object correctly and closer to the object frame.}\label{fig:manoeuvrer_phase}
\end{figure}

The reward in \eqref{eq:simple_rew} incentivizes the agent to approach the object along a straight trajectory, although it can be troublesome because the robots may collide or the GEN3 hand might be oriented with its back facing the object, as shown in Fig.~\ref{fig:manoeuvrer_phase}. For this reason, we restrict the reward to take the GEN3's end effector closer to the object than to the UR5e. Moreover, we propose a double frame system for the agent to maneuver into a suitable position for manipulation, with the GEN3 palm toward the target. If both conditions are satisfied, the agent is in a correct zone for approaching.
First, we calculate the distance between the palm of each robot $d_{\text{ROBOTS}_t}$. Then, we define a frame for the back of the GEN3 hand, computing the distance between it and the object $d_{\text{BACK}_t}$. In this way, we define a condition for the approach phase in \eqref{eq:mod_man}, which we will use in \eqref{eq:mod}.

\begin{equation} \label{eq:mod_man}
    m_{\text{MAN}_t} = (d_{\text{ROBOTS}_t} > d_t) \land (d_{\text{BACK}_t} > d_t)\,,
\end{equation}

where $m_{\text{MAN}_t} \in \{0, 1\}$ is a modifier for \eqref{eq:simple_rew} which we will use in the following equations.

\subsubsection{Approach phase}
Once the GEN3 palm is located in a suitable zone for approaching the object, the action taken at $t$ must bring the agent closer than it was at step $t-1$. Hence, in \eqref{eq:mod} we extend the maneuver modifier $m_{\text{MAN}_t}$ to restrict the distance at $t$ and so to be lower than the one in the previous step.

\begin{equation} \label{eq:mod}
    m_{t} = 2 [(d_t < d_{t-1}) \land m_{\text{MAN}_t}] - 1,
\end{equation}

where $(d_t < d_{t-1}) \in \{0, 1\}$ and $m_{t} \in \{-1, 1\}$.

\subsubsection{Handover phase}
\begin{figure}
    \centering
    \includegraphics[width=0.55\linewidth]{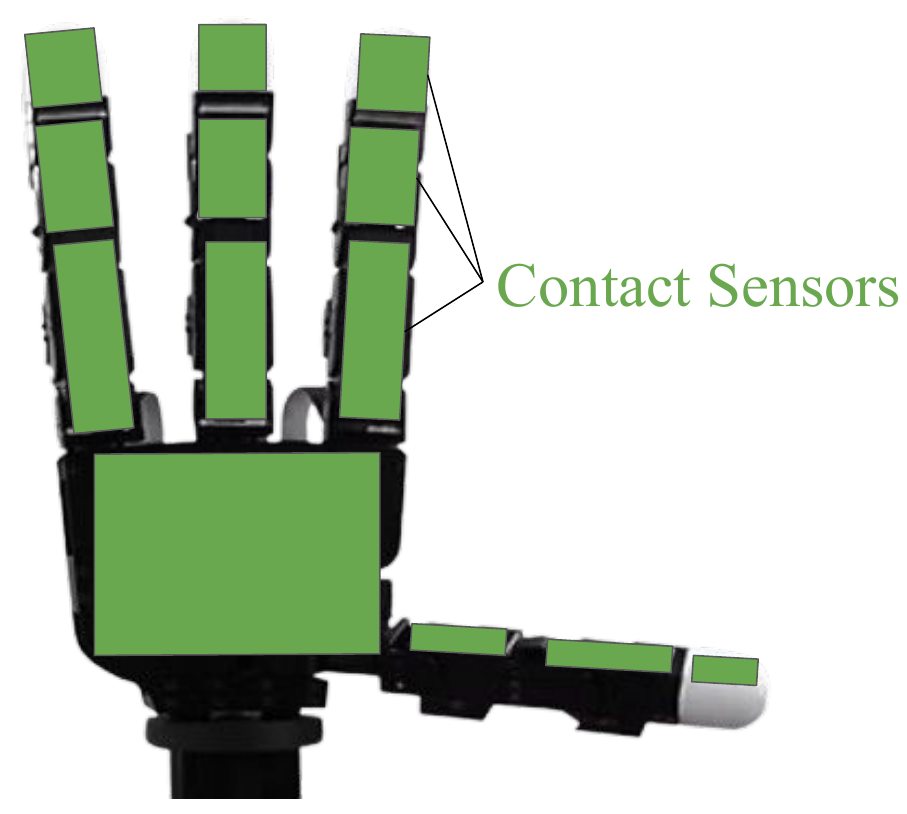}
    \caption{Placement of the contact sensors on the Allegro hand. Each green area represents a boolean contact sensor. It is important to note that contact sensing beyond the fingertips is necessary to successfully complete the object handover task as the object is mainly grasped with the palm and the proximal phalanges.}\label{fig:finger_sens}
\end{figure}

For the handover phase, we place tactile sensors on the phalanges of the fingers and the palm of the GEN3 hand, as shown in Fig.~\ref{fig:finger_sens}. In this way, we can detect the contacts between each finger and other elements of the simulation. We define the contacts as a vector of Boolean values $\vec c$ and we weight them using $\vec w_c$ according to their relevance in the task; the lower phalanges are of greater importance than the tips. In this work, it is worth mentioning that we only consider the contacts between the GEN3 hand and the object. In \eqref{eq:mod_scale} we modify the original weight $\eta_0 = 1$ according to the contacts. Therefore, the more touches, the lower that value will be.

 \begin{equation} \label{eq:mod_scale}
     \eta_t = \frac{\eta_0}{\sum_{0}^i \vec c_i + 1}\,.
 \end{equation}

In addition, we add the weighed sum of contacts $\vec c \cdot \vec w_c$ to the reward $r_{\text{BASE}_t}$. Thus, when the agent touches the object, the reward encourages it to close the hand rather than to continue advancing towards the target.

\subsubsection{Manipulation phase}
In the previous phases, the object is not grasped yet and the UR5e hand remains closed. This stage is changed when the thumb and another finger of the GEN3 hand touch the object. At this point, we consider the object susceptible to be grasped, so the UR5e hand opens. The reward also changes to take into account the distance between the object and the starting pose of the GEN3 $d_{\text{TGT}}$ following \eqref{eq:rew_manip}. Hence, the objective of this phase is to take the object to the starting pose of the GEN3.
\begin{equation} \label{eq:rew_manip}
    r_{\text{MANIP}_t} = \alpha \,  e^{-d_{\text{TGT}_t}},
\end{equation}
where $\alpha=12$ is the weight for the manipulation phase.
Moreover, when the agent changes to this phase or reaches the target, it receives a bonus.

The final reward function remain as \eqref{eq:final_reward}, involving all phases and modifiers.
\begin{equation} \label{eq:final_reward}
    r_t = \begin{dcases}
        m_{t} \,  r_{\text{BASE}_t} + \vec c \cdot \vec w_c&,  \text{ if} \;\; \text{\textbf{not}} \; (\mathbf{1}_{\text{G}}) \\
        r_{\text{MANIP}_t} + \vec c \cdot \vec w_c &, \text{ otherwise}
    \end{dcases} 
\end{equation}
where $\mathbf{1}_{\text{G}} \in \{0,1\}$ indicates whether the agent has already grasped the object or not.

\subsection{Distance calculation methods} \label{sec:distances}
We need to calculate the distances between poses in the environment to compute the rewards. When dealing with poses, it is important to define precise metrics that reflect not only translational distances but also orientation ones. For this reason, we use and compare several representations according to their formulation: dual quaternion, translation along with Euler angles, and homogeneous transformation matrices. 

\subsubsection{Dual Quaternions}
Given two dual quaternions $\mathbf{\hat q_1}$ and $\mathbf{\hat q_2}$, we compute the difference transformation between them, noted as $\mathbf{\hat q_{\text{diff}}}$, using the corresponding formula in Table \ref{tab:dq_operations}. 
This difference $\mathbf{\hat q_{\text{diff}}}$ is the identity element $\mathbf{\hat I}$ when they represent the same pose. Then, $\mathbf{\hat q_{\text{diff}}} - \mathbf{\hat I} \approx 0_{1\times8}$ if both transformations are the same. In this regard, we calculate the distance between frames using dual quaternions using \eqref{eq:dq_distance} as presented in \cite{velasco2025dualquat}. In this work, we propose this reward function to perform the object handover task.

\begin{equation} \label{eq:dq_distance}
    d_{\text{DQ}} = ||\mathbf{\hat q_{\text{diff}}} - \mathbf{\hat I}||_2,
\end{equation}

where $||\cdot||_2$ is the second norm of all elements of $\mathbf{\hat q_{\text{diff}}} - \mathbf{\hat I}$.

\subsubsection{Euler Angles}
Considering two sets of Euler angles $\vec e_1$ and $\vec e_2$ in the $XYZ$ convention with their corresponding Euclidean translations $\vec t_1$ and $\vec t_2$, we calculate the distance using this representation applying \eqref{eq:euler_distance}, extended from \cite{iriondo2023learning}.

\begin{equation} \label{eq:euler_distance}
    d_{\text{EULER}} = \psi||\vec t_1 - \vec t_2||_2 + \mu||\vec e_1 - \vec e_2||,
\end{equation}

where $\psi = 2.1$ and $\mu=0.32$ are scaling factors, so the magnitude of the distance in translation and rotation is similar due to differences in units.

\subsubsection{Homogeneous Transformation Matrices}
The poses are represented in homogeneous transformation matrices as
$\underline T_1$ and $\underline T_2$ being $\underline T = [R_{3\times3} \ \ \  \vec{t}; \ \  0_{1\times3} \ \ \   1]$, with $R_{3\times3} \in SO(3)$ as the rotation matrix and $\vec{t} \in \mathbb R^3$ as the Euclidean translation vector. Consequently, we obtain the distance by separating the translation and rotation parts from the elements as shown in \eqref{eq:euler_distance}. Therefore, we compute the orientation component using \eqref{eq:mat_angle}, which corresponds to the relative angle between the two rotation matrices.

\begin{equation} \label{eq:mat_angle}
    \theta_{\text{diff}} = \text{acos}(\frac{\text{trace}(R_{1_{3\times3}}^T \cdot R_{2_{3\times3}}) - 1}{2}),
\end{equation}

As a result, the final distance consists of the addition of the rotation and translation components, as shown in \eqref{eq:mat_distance}.

\begin{equation} \label{eq:mat_distance}
    d_{MAT} = \psi||\vec t_1 - \vec t_2||_2 + \beta \, \theta_{\text{diff}},
\end{equation}

where $\psi = 2.1$ and $\beta = 0.32$ are scaling factors to normalize the distances so the magnitudes are similar.

\section{Experiments}
\label{sec:experiments}
We now present several experiments to evaluate the performance of the agents produced by all the proposed reward functions, as well as to test the robustness of the system to novel objects. 
Specifically, we investigate the following questions:

\begin{itemize}
    \item Can we use \gls{rl} to train an object handover policy using two robotic arms equipped with multi-finger hands?
    \item How accurate and robust is the trained policy to seen and novel objects in simulation?
    \item Are dual quaternions more adequate to compute reward distances in $SE(3)$ than Euler or rotation matrices?
    \item Is the trained policy robust enough to handle motion perturbations during the object handover?
\end{itemize}

\subsection{Simulation}

We performed the entire training process along with experiments on the IsaacLab simulator \cite{mittal2023orbit} using the \gls{sb3} framework \cite{stable-baselines3}. In addition, we used \gls{ppo} \cite{schulman2017proximal} with three different seeds, using 1024 environments on a single NVIDIA A40 GPU. 
For each episode, we randomly reset the object position inside a cube of $\pm 0.15$ m of side, as well as the orientation with a variation of $\pm 0.3$ rad in roll and yaw, and $\pm 0.6$ rad in pitch.

\subsection{Learning to handover an object}

\begin{figure}[t]
    \centering
    \includegraphics[width=1.0\linewidth]{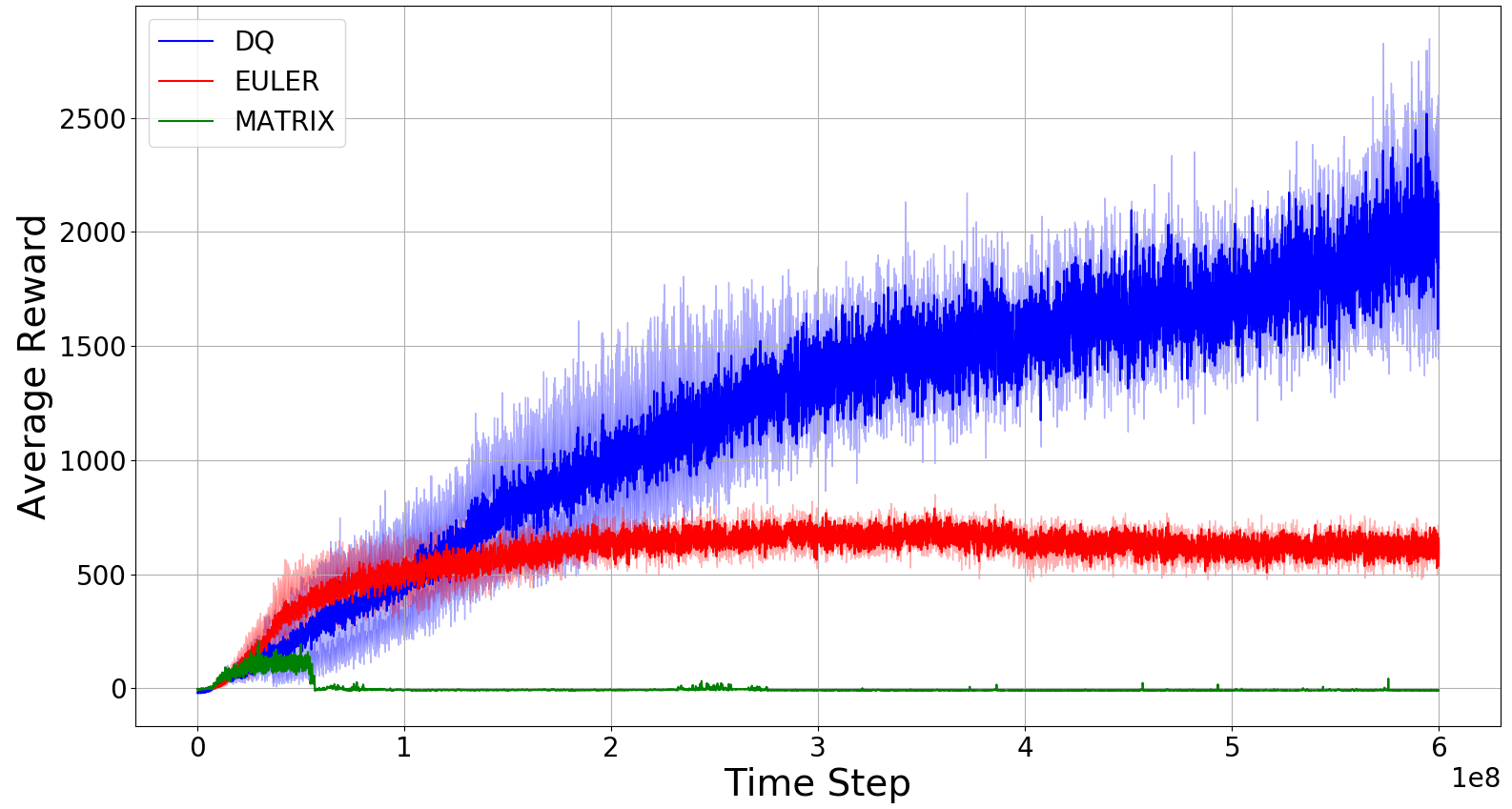}
    \caption{Average reward for all the agents trained for the object handover task using different reward functions: DQ (blue), EULER (red), MATRIX (green). The DQ agent shows the higher reward values, followed by the EULER and MATRIX agents.}
    \label{fig:train_plots}
\end{figure}

We chose an elongated quadrangular prism as the original training object of ($0.035 \times 0.035 \times 0.45$) m, as shown in Fig.~\ref{fig:env_setup}. The single object training aims to reduce computational resources and to assess how the policy performs under unseen objects in this setup. The training results in terms of the average reward are shown in Fig.~\ref{fig:train_plots}.

The MATRIX agent was unable to learn the task correctly, collapsing with a constant reward near zero. A possible reason for this issue may be the angle representation of the angle-vector to compute rotational distances. During training, there might be cases in which the observations were similar but the $\theta_{\text{diff}}$ for each observation was not. In these cases, the rotation vector was different as the angle remained the same. This can generate ambiguity when the policy obtained the actions. Thus, we did not consider this agent for the next experiments.

In contrast, the EULER agent learned faster to perform the task successfully compared to the DQ agent, but the reward became constant while the DQ agent continued to learn. Specifically, the EULER agent exploited the reward bonuses for grasping the object, thereby trying to reach those stages of the task quickly to increase the reward. However, this behavior led to an incorrect learning of the task during simulation, as explained in the following Section \ref{sec:success_rate_seen_novel}.

The DQ agent achieved the highest reward during the training process without stabilizing, indicating that the agent would collect an even higher reward if the training continued. The best results obtained by the DQ agent may be due to the ability of the proposed reward function to calculate the rotation differences or distances more accurately. 

Finally, we can answer the first question by saying that RL can be used to learn the dexterous object handover task. However, 
we observed that the results are highly dependent on the representation used to compute the rotation distances in the reward function, with Euler angles and dual quaternions being able to solve the task.

\subsection{Success rate with seen and novel objects}
\label{sec:success_rate_seen_novel}

\begin{figure}[t]
    \centering
    \subfigure[Success.]{\includegraphics[width=0.32\linewidth]{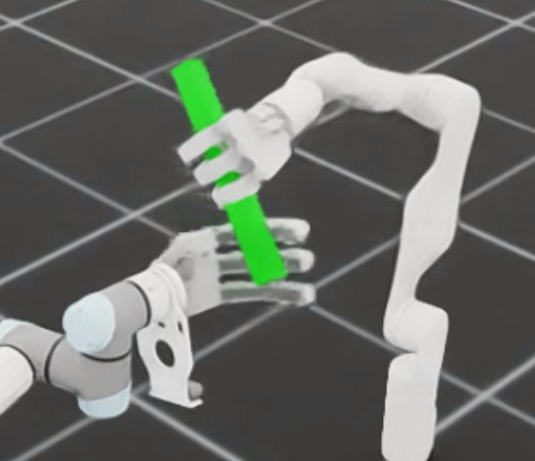}}
    \hfill
    \subfigure[Indetermination.]{\includegraphics[width=0.32\linewidth]{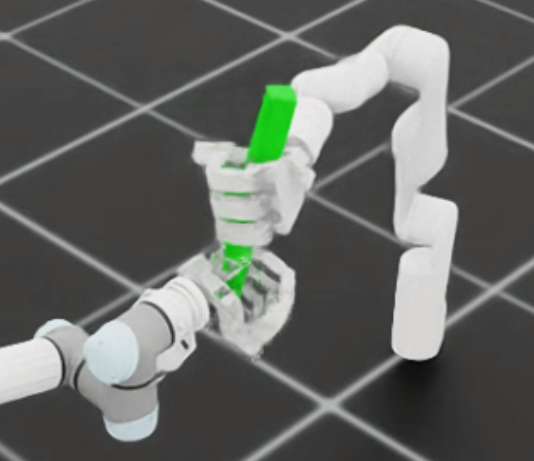}}
    \hfill
    \subfigure[Failure.]{\includegraphics[width=0.32\linewidth]{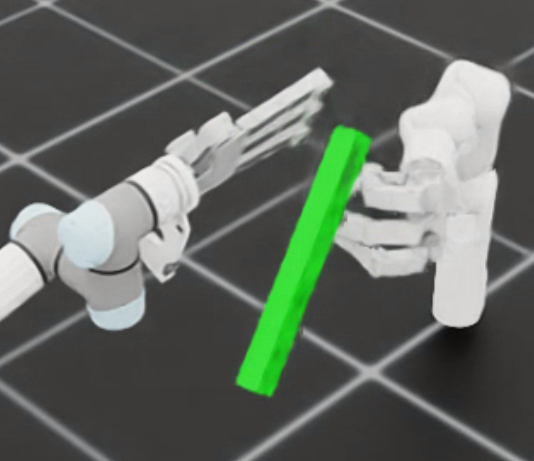}}

    \caption{Graphical description of the different cases contemplated during the experiments to evaluate the object handover. The success case is a correct handover without collision or falling. During the indetermination case, the object is clipped through the UR5e hand. The failure case happens when the object falls during handover.}
    \label{fig:exp_cases}
\end{figure}

We used the following metrics to measure the success rate during the experimentation (see Fig.~\ref{fig:exp_cases}).

\begin{itemize}
    \item \textbf{Success} (Succ.): the robot grasped and manipulated the object correctly.
    \item \textbf{Indetermination} (Ind.): indicates that, although the robot manipulated the object correctly, there are bugs during the episode because of failures of the simulator to resolve collisions. Therefore, in a real environment, we would not consider the grasp an absolute success.
    \item \textbf{Total Success} (Total Succ.): the combination of indetermination and success cases.
    \item \textbf{Fail}: indicates object falling during the episode.
\end{itemize}

We selected the best agents for each case to evaluate the trained policies using the same object as during training. The results in Table \ref{tab:succ_rate} show the performance for a total of 100 episodes in each case. The DQ agent achieved a total success rate of 91\%, while the EULER agent obtained 83\%. These results are an improvement of nearly 10\% from the DQ agent with respect to the EULER agent, although both have a considerable number of indeterminate cases. During tests, inconsistencies in the simulation may be causing movements that made the object clip through the UR5e hand, snatching the object out of its hand instead of grasping and taking it with subtlety. Furthermore, the EULER agent collided with the UR5e robot frequently, causing imprecisions during the episode that led to cases of indetermination.

\begin{table}[t]
\caption{Success rate of the agents expressed as a percentage (\%) when evaluating with the prism used during training. The DQ agent achieved the highest Total Success value compared to the EULER agent. The best results are in bold.} \label{tab:succ_rate}
\centering
\begin{tabular}{|c|cccc|}
\hline
Agent          & \multicolumn{1}{c}{Suc.} (\%) & Ind. (\%) & \begin{tabular}[c]{@{}c@{}}Total\\ Succ. (\%) \end{tabular} & Fail (\%) \\ \hline
\textbf{DQ}    & 59                      & 32 & \textbf{91}                                                  & 9 \\ \hline
\textbf{EULER} & 17                      & 66 & 83                                                  & 17 \\ \hline
\end{tabular}
\end{table}

The objective of the following experiments is to test the robustness of the agents to different types of unseen object morphologies. The new objects are a short prism of dimension ($0.035 \times 0.035 \times 0.35$) m, a cylinder of radius $0.019$ m and length of $0.45$ m, and a short cylinder with the same radius and a length of $0.35$ m. In Table \ref{tab:generalizab} the success rates for each agent with each object are shown.

\begin{table*}[t]
\centering
\caption{Success rate of the agents expressed as a percentage (\%) when evaluating with the objects out of the training distribution. Note that the DQ agent achieved the higher Total Sucess value compared to the EULER agent. The best results are in bold.} \label{tab:generalizab}
\resizebox{\textwidth}{!}{%

\begin{tabular}{c|cccc|cccc|cccc|}
\cline{2-13}
\textbf{}                            & \multicolumn{4}{c|}{\textbf{\begin{tabular}[c]{@{}c@{}}Short\\ Prism\end{tabular}}} & \multicolumn{4}{c|}{\textbf{Cylinder}}                                      & \multicolumn{4}{c|}{\textbf{\begin{tabular}[c]{@{}c@{}}Short\\ Cylinder\end{tabular}}} \\ \hline
\multicolumn{1}{|c|}{Agent}          & Succ. (\%)   & Ind. (\%)   & \begin{tabular}[c]{@{}c@{}}Total\\ Succ. (\%) \end{tabular}   & Fail (\%)  & Succ. (\%) & Ind. (\%) & \begin{tabular}[c]{@{}c@{}}Total\\ Succ. (\%) \end{tabular} & Fail (\%) & Succ. (\%)       & Ind. (\%)  & \begin{tabular}[c]{@{}c@{}}Total\\ Succ. (\%) \end{tabular}   & Fail (\%)  \\ \hline
\multicolumn{1}{|c|}{\textbf{DQ}}    & 69    & 25   & \textbf{94}                                                    & 6   & 41  & 46 & \textbf{87}                                                  & 13 & 55   &  35      & \textbf{90}                                                        &  10     \\ \hline
\multicolumn{1}{|c|}{\textbf{EULER}} & 32    & 60   & 92                                                    & 8   & 8  & 76 & 84                                                  & 16 &    25         &  61      &                                                     86    & 14      \\ \hline
\end{tabular}}
\end{table*}

When testing with the short prism, the DQ and EULER agents obtained a total success rate of 94\% and 92\%, respectively, which are higher rates with respect to the previous experiment. 
The smaller size of the object may be causing a more precise and stable grasp because the robot is approaching better the grasp frame. Although they have similar total success rates, the EULER agent obtained many more cases of indetermination owing to the same reasons as when testing with the original prism.
Regarding the results with respect to the cylinder objects in Table \ref{tab:generalizab}, the circular shape of its surface can cause the orientation of the object to slightly vary inside the UR5e hand. This rotates the target frame to unseen poses, leading to a lower success rate for all agents with 87\% and 84\% for the DQ and EULER. When testing with the shorter version of the cylinder, a success rate of 90\% and 86\% is obtained for the DQ and EULER agent, respectively. It is relevant to consider that both agents obtained around 20\% less indetermination cases with the shorter version of the cylinder. 

Therefore, to respond to the second question, we can confirm that the trained policy is accurate when evaluating with objects from the training distribution and is also robust to objects from another distribution. Specifically, the best results are achieved when manipulating short rectangular objects.

\subsection{Distance minimization with dual quaternions}

\begin{figure*}[t]
    \centering
    \subfigure[Global distance in Dual Quaternions for the DQ agent.]{\includegraphics[width=0.32\textwidth]{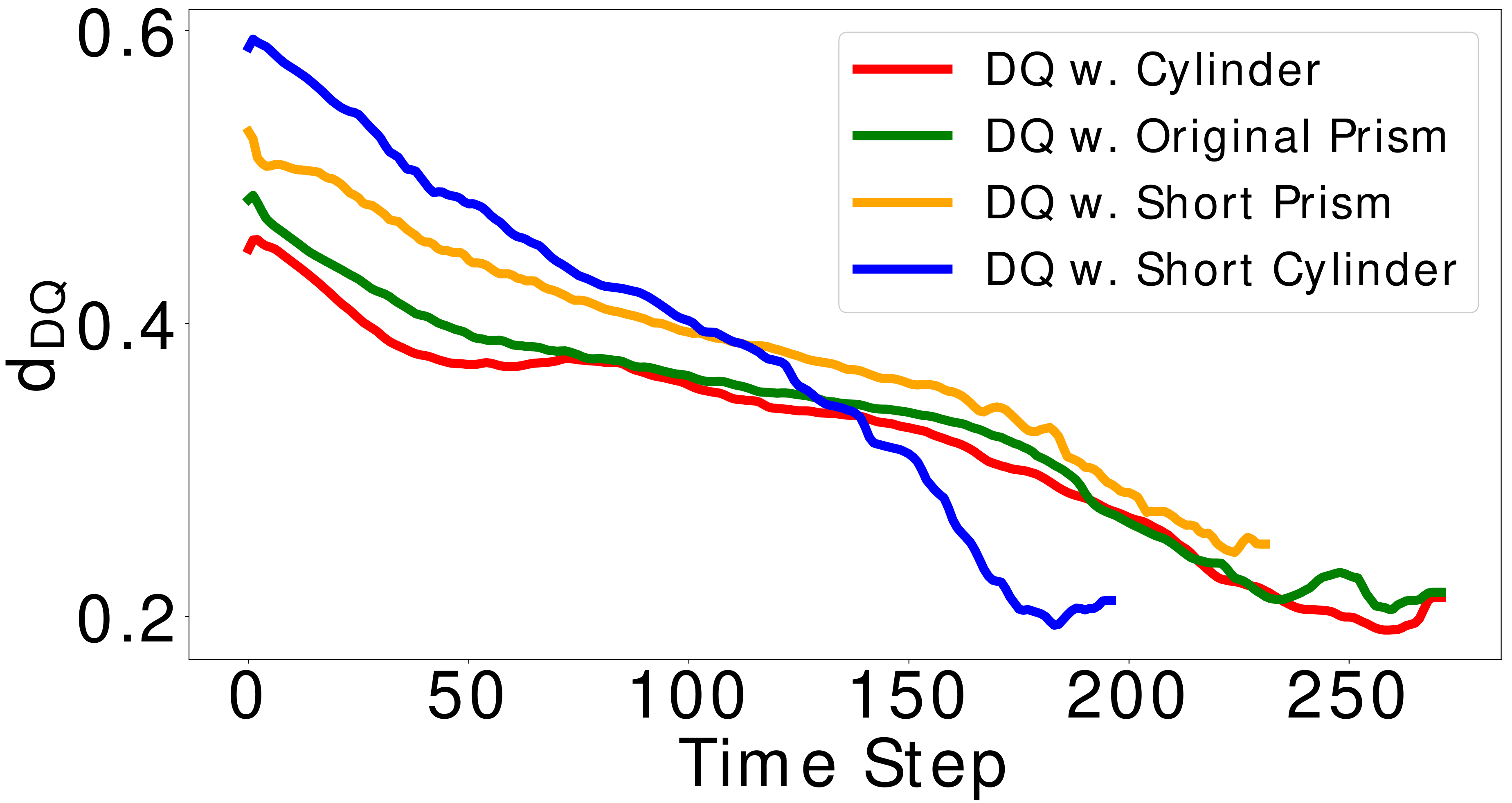}}\hfill
    \subfigure[Euclidean Translation for the DQ agent.]{\includegraphics[width=0.32\textwidth]{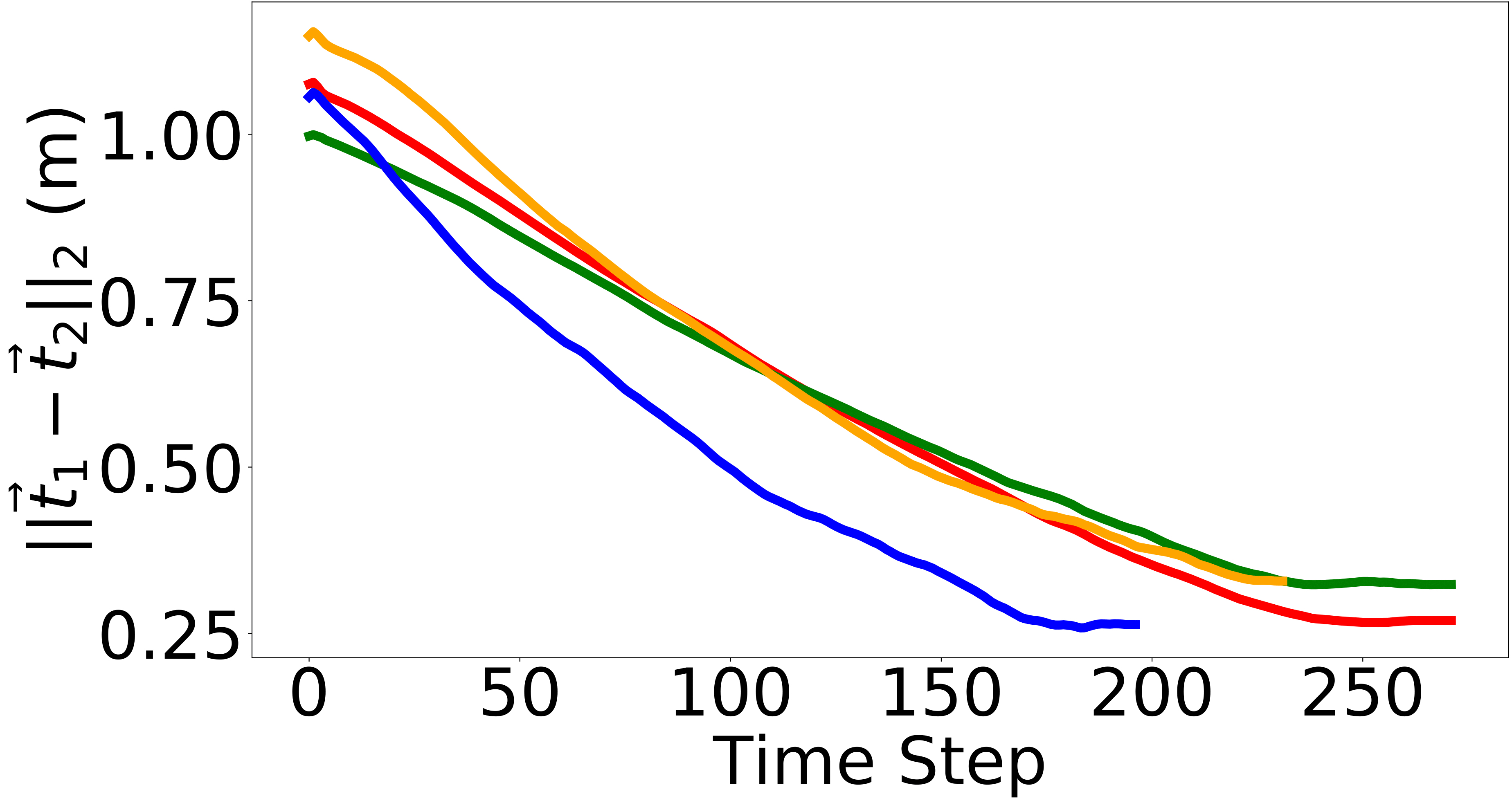}}\hfill
    \subfigure[Rotation Distance in Dual Quaternions for the DQ agent.]{\includegraphics[width=0.32\textwidth]{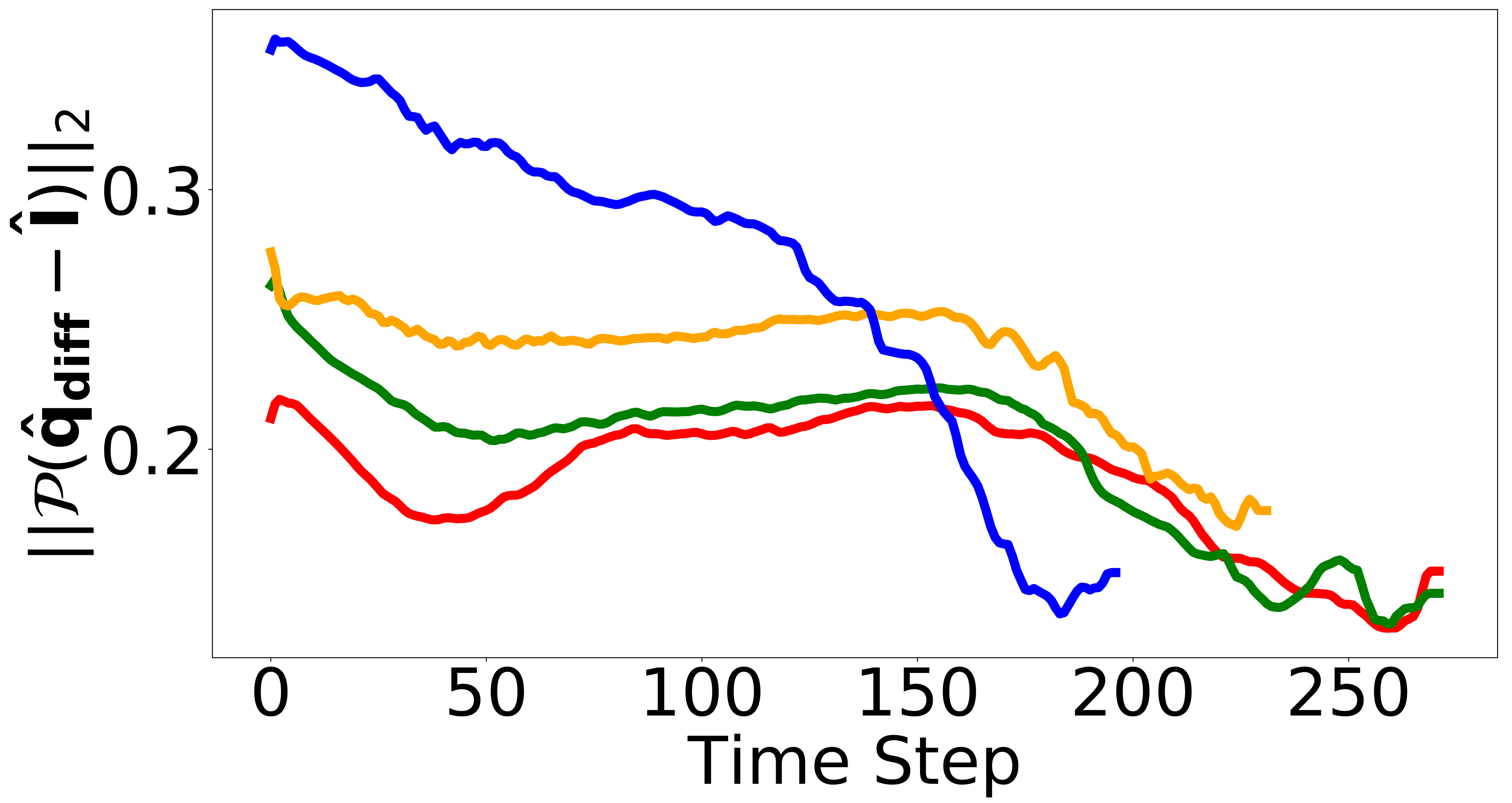}}
    \subfigure[Global distance in Translation + Euler for the EULER agent.]{\includegraphics[width=0.32\textwidth]{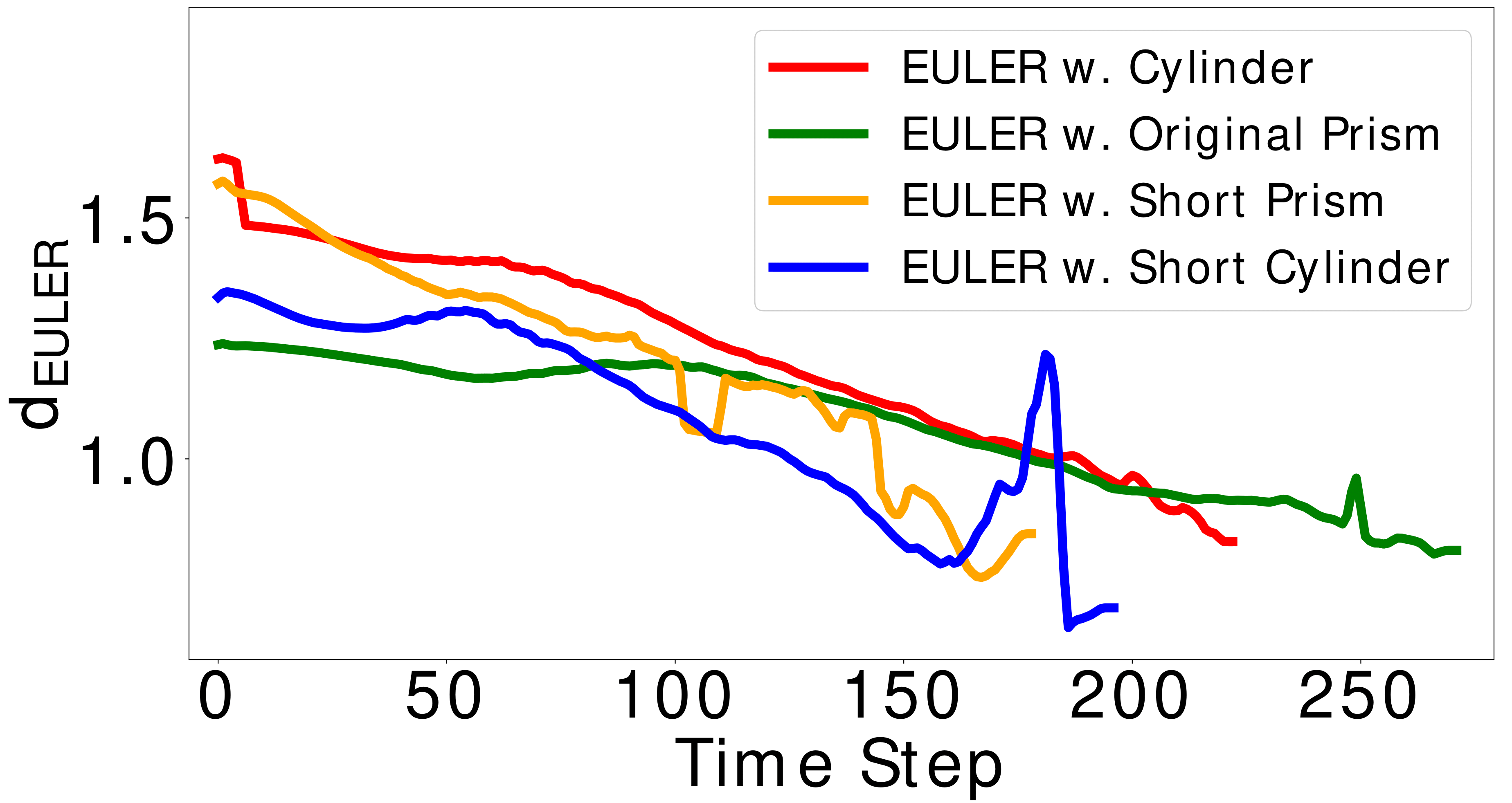}}\hfill
    \subfigure[Euclidean Translation for the EULER agent.]{\includegraphics[width=0.32\textwidth]{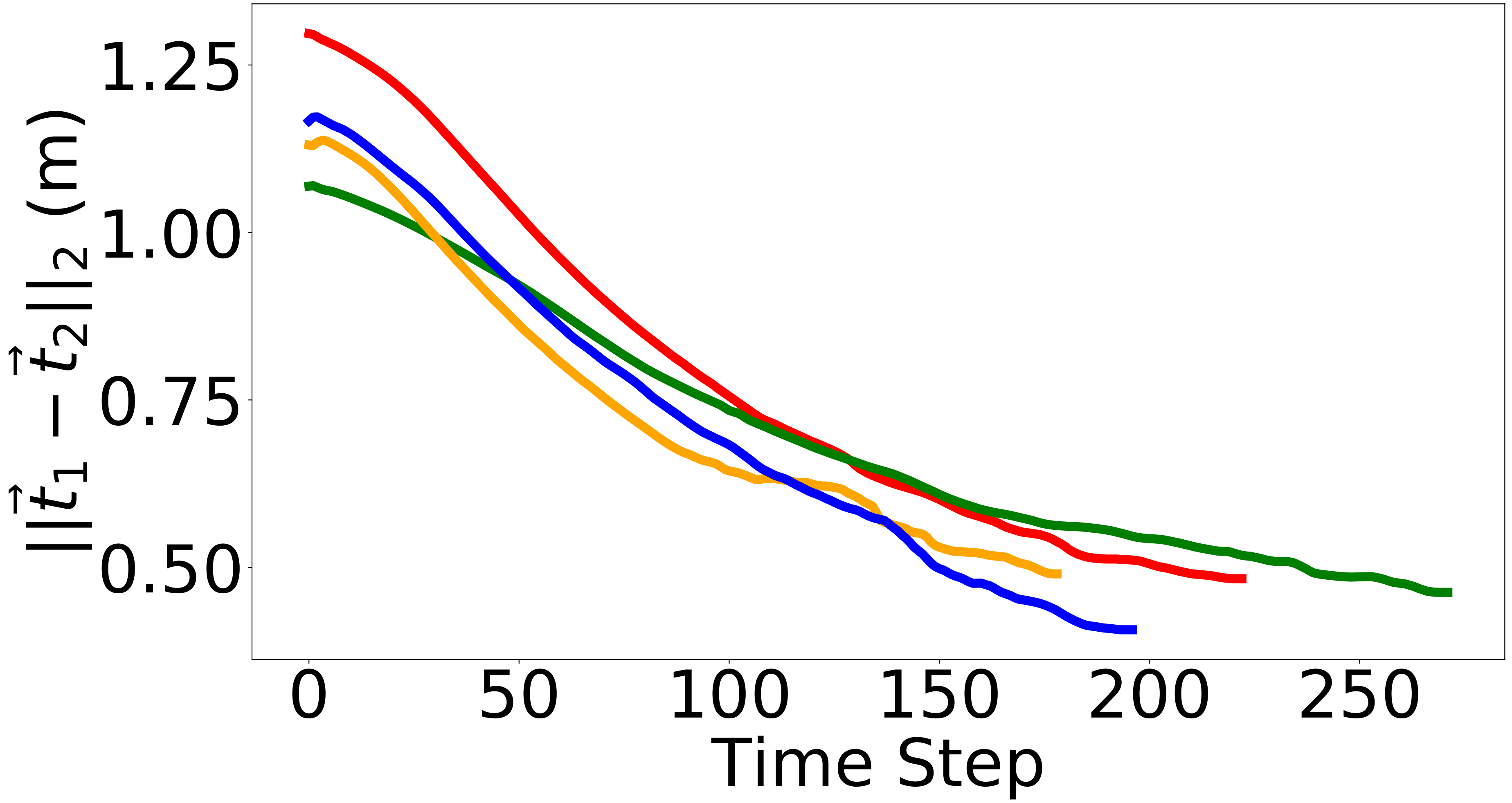}}\hfill
    \subfigure[Rotation Distance for the EULER agent.]{\includegraphics[width=0.32\textwidth]{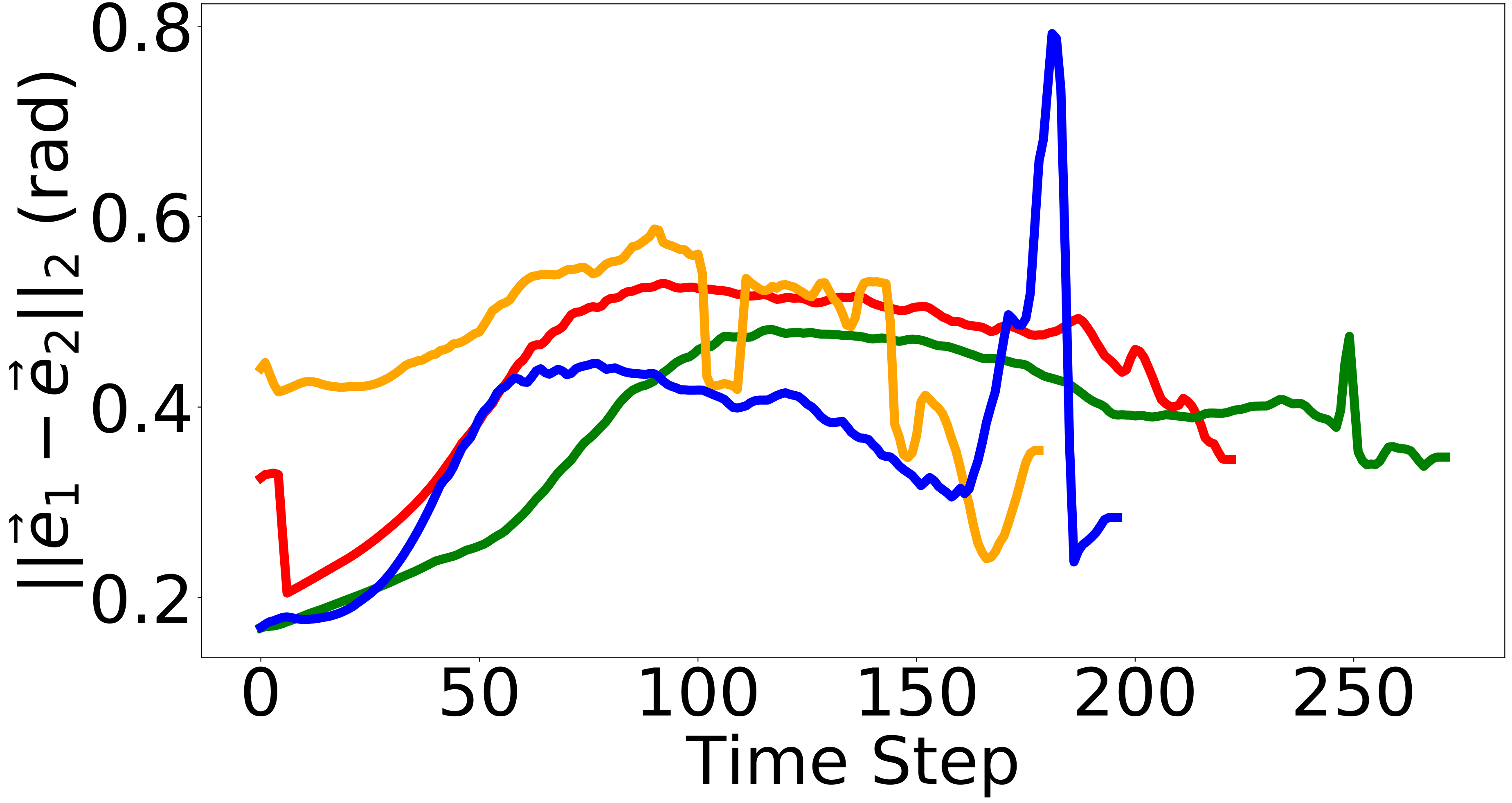}}
    \caption{Distance minimization of the agents for all selected objects. The DQ agent showed a greater minimization of the rotation distance than the EULER agent, which maximized it in most cases.}
    \label{fig:dist_min}
\end{figure*}

The objective of these experiments is to test how precise the trajectory followed by the agent is from the starting point to the target frame of the object. 
We conducted a total of 10 successful experiments for each agent with the different objects proposed. During those, we collected how the distance to the object frame is minimized. In Fig.~\ref{fig:dist_min}, all three minimization cases are shown for both agents, confirming the results in Tables \ref{tab:succ_rate} and \ref{tab:generalizab}.

For the DQ agent, we calculated the mean distances following \eqref{eq:dq_distance} in each step until the target pose is reached. Given two frames in space, we obtained the translation distance by applying $||\vec t_1 - \vec t_2||_2$, which represents the Euclidean distance between two poses. On the other hand, we calculated the rotational distance using $||\mathcal{P}(\mathbf{\hat q_{\text{diff}}} - \mathbf{\hat I})||_2$, which represents the rotation term of the dual quaternion difference of both poses. Alternatively, we computed the global distance for the EULER agent using \eqref{eq:euler_distance}, while we extracted the translational and rotational distances from the respective terms of the addition in that equation.

Both policies minimized the global distance from the target (Fig.~\ref{fig:dist_min}a and \ref{fig:dist_min}d). However, the DQ one made it smoother than the EULER.  Even though both agents were able to minimize the translational distance (Fig.~\ref{fig:dist_min}b and \ref{fig:dist_min}e), the EULER one failed to orient the end effector to the target rotation as shown in Fig.~\ref{fig:dist_min}f. The DQ agent minimized the rotational poses from the starting pose around 43\% on average. Far from reducing it, the EULER one increased it. This fact may be one of the reasons why the latter obtains more indetermination cases during the tests. 

Answering the third question, we state that dual quaternions are more adequate to calculate the rewards on $SE(3)$ for approach trajectories. The distance minimized by the agents trained with this representation is lower and smoother than the obtained using the EULER representation.

\subsection{Success rate under perturbations}
When dealing with object handover in the real world, it is common to have perturbations on the object pose from the giver agent. For this reason, we conducted some experiments where the handing robot (UR5e) is moving during the object handover. The objective of these is to test the robustness of the policy to unseen states where there is movement of the target.
In these experiments, the UR5e is moving in a random direction within the limits where the agent was trained. The velocity is set to be 40\% lower than GEN3 so it can reach it. Specifically, at $0.03$ m/s for the linear velocity and $0.16$ rad/s for the angular velocity. The results are shown in Table~\ref{tab:res_pert}. All the total success rates decreased around 13.68\% on average due to movement during the task in our setup.
The best results are still achieved by the DQ agent when manipulating the original or short prism with 81\% success compared to a 78\% from the EULER. The DQ agent performed better with all objects except the cylinder, where the EULER obtained 77\% compared to a 71\% of total success from the DQ.
The movement of the UR5e caused the agent to lose precision in the task, although it still tried to perform the handover even taking advantage of the simulation problems.

\begin{table*}[t]
\caption{Success rate of the agents expressed as a percentage (\%) when evaluating the trained policies under perturbations. The DQ agent showed higher Total Success values for 3 out of the 4 objects compared to the EULER one. The best results are in bold.} \label{tab:res_pert}
\resizebox{\textwidth}{!}{%
\centering
\Huge
\begin{tabular}{c|cccc|cccc|cccc|cccc|}
\cline{2-17}
\textbf{}                            & \multicolumn{4}{c|}{\textbf{Prism}}                                         & \multicolumn{4}{c|}{\textbf{\begin{tabular}[c]{@{}c@{}}Short\\ Prism\end{tabular}}} & \multicolumn{4}{c|}{\textbf{Cylinder}}                                      & \multicolumn{4}{c|}{\textbf{\begin{tabular}[c]{@{}c@{}}Short\\ Cylinder\end{tabular}}} \\ \hline
\multicolumn{1}{|c|}{Agent}          & Succ. (\%)   & Ind. (\%)   & \begin{tabular}[c]{@{}c@{}}Total\\ Succ. (\%)  \end{tabular} & Fail (\%)   & Succ. (\%)    & Ind. (\%)    & \begin{tabular}[c]{@{}c@{}}Total\\ Succ. (\%) \end{tabular}   & Fail (\%)    & Succ. (\%)   & Ind. (\%)   & \begin{tabular}[c]{@{}c@{}}Total\\ Succ. (\%)  \end{tabular} & Fail (\%)  & Succ. (\%)      & Ind. (\%)      & \begin{tabular}[c]{@{}c@{}}Total \\ Succ. (\%)  \end{tabular}    & Fail (\%)    \\ \hline
\multicolumn{1}{|c|}{\textbf{DQ}}    &   46    &  35    & \textbf{81}                                                      &   19   &  55  & 26   & \textbf{81}                                          &  19   & 37  & 34 & 71                                        & 29 &  12    &  65   &  \textbf{77}  & 23   \\ \hline
\multicolumn{1}{|c|}{\textbf{EULER}} &   11    & 67     & 78                                                      &  22    &  21   &  51  & 72                                                    &  28  & 12  & 65 & \textbf{77}                                                  & 23 & 13     & 59    & 72                                                     &  28  \\ \hline
\end{tabular}}
\end{table*}

As a result and answering the fourth question, our policies are robust to movement of the object during the handover. In particular, the DQ agent obtained a higher success rate with almost all proposed objects.

\section{Conclusions}
\label{sec:conclusions}

In this work, we evaluate in simulation an \gls{rl} policy to perform dexterous object handover with a multi-finger hand. We trained the policy by using the PPO algorithm, a single quadrangular prism as the handover object, and a variety of random poses to reset the object differently at the beginning of each episode. 
Experimental results demonstrate that the policy trained with the quadrangular prism object is robust to other objects with similar geometric shapes and also with different sizes. Furthermore, it is shown that the dual quaternion representation can minimize the orientation of the hand with respect to the target orientation of the object, while other representations failed. Finally, the trained policy was evaluated under perturbations during the object handover, showing only minimal decrease in performance after the same number of experiments.

While this work is a promising first step towards dexterous object handover, it is limited by several factors: First, a single object is employed for the training process, thereby diminishing the generalization capabilities of the policy. Second, the proposed policy does not incorporate any kind of visual perception; it is only extracting the object pose from the simulator. This is very important to transfer the trained policy from simulation to the real-world setup. Finally, imperfections in the physics engine limit the training of the policy by enabling undesired grasping behaviors that may not occur in the real-world setup.
Future work will look into training not only the receiving robot, but also the handing robot within the RL framework used, as well as looking into the generalization of the policy to a wider set of objects. 

Moreover, it is planned to transfer the policy to real-world scenarios. In this way, it could be assessed how the policy behaves under noisy observations provided by sensors in real environments. By doing so, it will remove the indetermination cases present during the simulation and test the trained policies with humans as givers.

\bibliographystyle{IEEEtranDOI}
\bibliography{IEEEabrv,refs}

\begin{thebibliography}{10}
\providecommand{\doi}[1]{doi: #1}
\providecommand{\url}[1]{#1}
\csname url@samestyle\endcsname
\providecommand{\newblock}{\relax}
\providecommand{\bibinfo}[2]{#2}
\providecommand{\BIBentrySTDinterwordspacing}{\spaceskip=0pt\relax}
\providecommand{\BIBentryALTinterwordstretchfactor}{4}
\providecommand{\BIBentryALTinterwordspacing}{\spaceskip=\fontdimen2\font plus
\BIBentryALTinterwordstretchfactor\fontdimen3\font minus \fontdimen4\font\relax}
\providecommand{\BIBforeignlanguage}[2]{{%
\expandafter\ifx\csname l@#1\endcsname\relax
\typeout{** WARNING: IEEEtran.bst: No hyphenation pattern has been}%
\typeout{** loaded for the language `#1'. Using the pattern for}%
\typeout{** the default language instead.}%
\else
\language=\csname l@#1\endcsname
\fi
#2}}
\providecommand{\BIBdecl}{\relax}
\BIBdecl

\bibitem{survey_object_handover}
V.~Ortenzi, A.~Cosgun, T.~Pardi, W.~P. Chan, E.~Croft, and D.~Kuli{\'c}, ``Object handovers: A review for robotics,'' \emph{IEEE Transactions on Robotics}, vol.~37, no.~6, pp. 1855--1873, 2021.  \href{http://dx.doi.org/\doi{10.1109/TRO.2021.3075365}}{\doi{10.1109/TRO.2021.3075365}}

\bibitem{DUAN2024100145}
D.~Haonan, Y.~Yifan, L.~Daheng, and W.~Peng, ``Human–robot object handover: Recent progress and future direction,'' \emph{Biomimetic Intelligence and Robotics}, vol.~4, no.~1, p. 100145, 2024.  \href{http://dx.doi.org/\doi{10.1016/j.birob.2024.100145}}{\doi{10.1016/j.birob.2024.100145}}

\bibitem{intro_human_robot_colab}
M.~Costanzo, G.~D. Maria, and C.~Natale, ``Handover control for human-robot and robot-robot collaboration,'' \emph{Frontiers in Robotics and AI}, vol.~8, 2021.  \href{http://dx.doi.org/\doi{10.3389/frobt.2021.672995}}{\doi{10.3389/frobt.2021.672995}}

\bibitem{rel_work_bim_hand_2}
Y.~Li, C.~Pan, H.~Xu, X.~Wang, and Y.~Wu, ``Efficient bimanual handover and rearrangement via symmetry-aware actor-critic learning,'' in \emph{IEEE International Conference on Robotics and Automation (ICRA)}, 2023.  \href{http://dx.doi.org/\doi{10.1109/ICRA48891.2023.10160739}}{\doi{10.1109/ICRA48891.2023.10160739}}

\bibitem{bimanual_1}
B.~Huang, Y.~Wang, X.~Yang, Y.~Luo, and Y.~Li, ``3d-vitac: Learning fine-grained manipulation with visuo-tactile sensing,'' in \emph{8th Annual Conference on Robot Learning}, 2024.  \href{http://dx.doi.org/\doi{10.48550/arXiv.2410.24091}}{\doi{10.48550/arXiv.2410.24091}}

\bibitem{10657347}
Z.~Wang, J.~Chen, Z.~Chen, P.~Xie, R.~Chen, and L.~Yi, ``Genh2r: Learning generalizable human-to-robot handover via scalable simulation, demonstration, and imitation,'' in \emph{IEEE/CVF Conference on Computer Vision and Pattern Recognition (CVPR)}, 2024, pp. 16\,362--16\,372.  \href{http://dx.doi.org/\doi{10.1109/CVPR52733.2024.01548}}{\doi{10.1109/CVPR52733.2024.01548}}

\bibitem{trabajo_prev_dani}
D.~Frau-Alfaro, S.~T. Puente, I.~De~Loyola Páez-Ubieta, and E.~Velasco-Sánchez, ``Robotic approach trajectory using reinforcement learning with dual quaternions,'' in \emph{7th Iberian Robotics Conference (ROBOT)}, 2024.  \href{http://dx.doi.org/\doi{10.1109/ROBOT61475.2024.10796878}}{\doi{10.1109/ROBOT61475.2024.10796878}}

\bibitem{intro_human_robot_colab2}
W.~He, J.~Li, Z.~Yan, and F.~Chen, ``Bidirectional human–robot bimanual handover of big planar object with vertical posture,'' \emph{IEEE Transactions on Automation Science and Engineering}, pp. 1180--1191, 2022.  \href{http://dx.doi.org/\doi{10.1109/TASE.2020.3043480}}{\doi{10.1109/TASE.2020.3043480}}

\bibitem{intro_human_robot_colab3}
S.~E. Ovur and Y.~Demiris, ``Naturalistic robot-to-human bimanual handover in complex environments through multi-sensor fusion,'' \emph{IEEE Transactions on Automation Science and Engineering}, pp. 3730--3741, 2024.  \href{http://dx.doi.org/\doi{10.1109/TASE.2023.3284668}}{\doi{10.1109/TASE.2023.3284668}}

\bibitem{rel_work_bim_hand_1}
B.~Huang, Y.~Chen, T.~Wang, Y.~Qin, Y.~Yang, N.~Atanasov, and X.~Wang, ``Dynamic handover: Throw and catch with bimanual hands,'' in \emph{7th Annual Conference on Robot Learning}, 2023.  \href{http://dx.doi.org/\doi{10.48550/arXiv.2309.05655}}{\doi{10.48550/arXiv.2309.05655}}

\bibitem{rel_work_bim_hand_3}
T.~Lin, K.~Sachdev, L.~Fan, J.~Malik, and Y.~Zhu, ``Sim-to-real reinforcement learning for vision-based dexterous manipulation on humanoids,'' \emph{arXiv preprint arXiv:2502.20396}, 2025.  \href{http://dx.doi.org/\doi{10.48550/arXiv.2502.20396}}{\doi{10.48550/arXiv.2502.20396}}

\bibitem{li2017deep}
Y.~Li, ``Deep reinforcement learning: An overview,'' \emph{arXiv preprint arXiv:1701.07274}, 2017.  \href{http://dx.doi.org/\doi{10.48550/arXiv.2412.05265}}{\doi{10.48550/arXiv.2412.05265}}

\bibitem{wiering2012reinforcement}
M.~A. Wiering and M.~Van~Otterlo, ``Reinforcement learning,'' \emph{Adaptation, learning, and optimization}, vol.~12, no.~3, p. 729, 2012.

\bibitem{jia2013dual}
\BIBentryALTinterwordspacing
Y.-B. Jia, ``Dual quaternions,'' \emph{Iowa State University: Ames, IA, USA}, 2013. [Online]. Available: \url{https://faculty.sites.iastate.edu/jia/files/inline-files/dual-quaternion.pdf}
\BIBentrySTDinterwordspacing

\bibitem{thomas2014approaching}
F.~Thomas, ``Approaching dual quaternions from matrix algebra,'' \emph{IEEE Transactions on Robotics}, 2014.  \href{http://dx.doi.org/\doi{10.1109/TRO.2014.2341312}}{\doi{10.1109/TRO.2014.2341312}}

\bibitem{velasco2025dualquat}
E.~P. Velasco-Sánchez, L.~F. Recalde, G.~Li, F.~A. Candelas-Herias, S.~T. Puente-Mendez, and F.~Torres-Medina, ``Dualquat-loam: Lidar odometry and mapping parameterized on dual quaternions,'' \emph{Robotics and Autonomous Systems}, 2025.  \href{http://dx.doi.org/\doi{10.1016/j.robot.2025.105009}}{\doi{10.1016/j.robot.2025.105009}}

\bibitem{iriondo2023learning}
A.~Iriondo, E.~Lazkano, A.~Ansuategi, A.~Rivera, I.~Lluvia, and C.~Tub{\'\i}o, ``Learning positioning policies for mobile manipulation operations with deep reinforcement learning,'' \emph{International journal of machine learning and cybernetics}, 2023.  \href{http://dx.doi.org/\doi{10.1007/s13042-023-01815-8}}{\doi{10.1007/s13042-023-01815-8}}

\bibitem{mittal2023orbit}
M.~Mittal, C.~Yu, Q.~Yu, J.~Liu, N.~Rudin, D.~Hoeller, J.~L. Yuan, R.~Singh, Y.~Guo, H.~Mazhar, A.~Mandlekar, B.~Babich, G.~State, M.~Hutter, and A.~Garg, ``Orbit: A unified simulation framework for interactive robot learning environments,'' \emph{IEEE Robotics and Automation Letters}, 2023.  \href{http://dx.doi.org/\doi{10.1109/LRA.2023.3270034}}{\doi{10.1109/LRA.2023.3270034}}

\bibitem{stable-baselines3}
\BIBentryALTinterwordspacing
A.~Raffin, A.~Hill, A.~Gleave, A.~Kanervisto, M.~Ernestus, and N.~Dormann, ``Stable-baselines3: Reliable reinforcement learning implementations,'' \emph{Journal of Machine Learning Research}. [Online]. Available: \url{http://jmlr.org/papers/v22/20-1364.html}
\BIBentrySTDinterwordspacing

\bibitem{schulman2017proximal}
J.~Schulman, F.~Wolski, P.~Dhariwal, A.~Radford, and O.~Klimov, ``Proximal policy optimization algorithms,'' \emph{arXiv preprint arXiv:1707.06347}, 2017.  \href{http://dx.doi.org/\doi{10.48550/arXiv.1707.06347}}{\doi{10.48550/arXiv.1707.06347}}

\end{thebibliography}

\end{document}